\documentclass[runningheads]{llncs}
\usepackage[noend]{algpseudocode}
\usepackage{algorithm}
\usepackage{amsmath}
\usepackage{amssymb}
\usepackage{todonotes}
\usepackage[frozencache,cachedir=.]{minted}

\usepackage{hyperref}
\usepackage{booktabs}
\usepackage[misc]{ifsym}

\usepackage[T1]{fontenc}
\usepackage{graphicx}
\begin{document}
\title{cuSLINK: Single-linkage Agglomerative Clustering on the GPU}

\author{Corey J. Nolet\inst{1,2} \Letter \thanks{Fist published in ECML PKDD 2023 by Springer Nature} \and  %
Divye Gala\inst{1}  \and  %
Alex Fender\inst{1} \and  %
Mahesh Doijade\inst{1} \and  %
Joe Eaton\inst{1} \and 
Edward Raff\inst{2,3} \and
John Zedlewski\inst{1} \and
Brad Rees\inst{1} \and
Tim Oates\inst{2}}
\authorrunning{C.J. Nolet et al.}
\institute{NVIDIA, Inc, Santa Clara, CA, USA \\ \email{\{cnolet,dgala,afender,mdoijade,featon,brees,jzedlewski\}@nvidia.com}\and
University of Maryland, Baltimore County, Baltimore, MD, USA \\\email{oates@umbc.edu}\and
Booz Allen Hamilton, McLean, VA, USA \\
\email{raff\_edward@bah.com}}

\titlerunning{cuSLINK: Single-linkage Agglomerative Clustering on the GPU}

\toctitle{cuSLINK: Single-linkage Agglomerative Clustering on the GPU}

\tocauthor{Corey~J.~Nolet, Divye~Gala, Alex~Fender, Mahesh~Doijade, Joe~Eaton, Edward~Raff, John~Zedlewski, Brad~Rees, Tim~}

\maketitle
\begin{abstract}
In this paper, we propose cuSLINK, a novel and state-of-the-art reformulation of the SLINK algorithm on the GPU which requires only $O(Nk)$ space and uses a parameter $k$ to trade off space and time. We also propose a set of novel and reusable building blocks that compose cuSLINK. These building blocks include highly optimized computational patterns for $k$-NN graph construction, spanning trees, and dendrogram cluster extraction. We show how we used our primitives to implement cuSLINK end-to-end on the GPU, further enabling a wide range of real-world data mining and machine learning applications that were once intractable. In addition to being a primary computational bottleneck in the popular HDBSCAN algorithm, the impact of our end-to-end cuSLINK algorithm spans a large range of important applications, including cluster analysis in social and computer networks, natural language processing, and computer vision. Users can obtain cuSLINK at \url{https://docs.rapids.ai/api/cuml/latest/api/#agglomerative-clustering}


\end{abstract}

\section{Introduction}

Hierarchical agglomerative clustering (HAC) is an important and fundamental algorithm for classical machine learning and data mining. HAC variants are used in many different informatics disciplines such as micro array analysis, genome clustering, computer vision, document clustering, and social network analysis \cite{Shalom2010}. In particular, the Single-Link HAC is still critically important in bioinformatics and genomics\cite{pmid36224396,pmid36191000,pmid30612741}, but a challenge to use due to its $\mathcal{O}(N^2)$ cost where $N$ is the number of items to be clustered. Hierarchical agglomerative clustering builds up a hierarchy of clusters from a set of vectors bottom-up, by starting with each vector in its own cluster and merging pairs of vectors together based on predefined similarity criteria until a common root is reached, which produces this quadratic lower bound.

The seminal SLINK algorithm~\cite{sibson_slink} provided the first $\mathcal{O}(N^2)$ time and $\mathcal{O}(N)$ space algorithm for the Single-link HAC problem. However, it has evaded efficient parallel implementation due to a lack of bulk work in the SLINK algorithm, causing communication overhead to dominate runtime. In this work, we will instead use a Minimum Spanning Tree (MST) based approach to the Single-link HAC that allows us to perform more parallel work, but at a potentially larger compute and memory complexity of $\mathcal{O}(N^2 + N k \log N)$ and $\mathcal{O}(N k)$ respectively. This required inventing a new parallel MST approach that efficiently performs the distance computations by an iterative expansion of a $k$-nearest neighbor graph. We find in practice, this trade-off is worth it by enabling a GPU implementation capable of 2290$\times$ faster runtime over common CPU implementations, and in all our tests is faster than alternative options today. An added benefit of this primitives-based approach is the ability to generalize cuSLINK for other distance measures, even in non-metric spaces.

We term our approach and implementation cuSLINK for its use of CUDA APIs and it produces an exact solution to the SLHC problem. The primitives we used to build cuSLINK are modular and can be reused to compose other important algorithms for graph and machine learning. For this reason, we provide separate benchmarks for these primitives, in addition to end-to-end benchmarks of our cuSLINK implementation against the currently available state-of-the-art. cuSLINK and its primitives are fully open source and have been contributed upstream to the RAFT library (https://github.com/rapidsai/raft).

In the following \autoref{sec:related}, we will outline related works and better shape the motivation for both our modular design and contributions. We present our contribution in \autoref{sec:description}, and detail our primitives, as well as our novel reformulation of single-linkage hierarchical clustering for the GPU. These primitives include constructing knn graphs by fusing the k-selection operation with the computation of distances, computing an MST using a variant of Borůvka's classic parallel algorithm, and a novel method for extracting flattened clusters from a dendrogram. \autoref{sec:benchmarks} compares the performance of our single-linkage hierarchical clustering algorithm, as well as its building blocks, against previous works.

\section{Related Work}
\label{sec:related}

Single-linkage hierarchical clustering is commonly referred to as 'nearest neighbors' clustering and as such, our implementation makes direct use of nearest neighbors computations in order to shrink the memory footprint of the naive computation of single-linkage clustering and providing a GPU-accelerated version that borrows its foundation from the original SLINK \cite{sibson_slink}.

The original SLINK algorithm maintains 3 arrays, each of size $n$ and loops over the range $[1, n]$, building up the dendrogram level by level, using two arrays of pointers to represent the dendrogram itself and a third to store distances. Sibsen notes that $O(n^2)$ is the optimally efficient runtime upper bound because each neighbor pair will ultimately need to be compared at least once. The challenge with this algorithm, however, is in SLINK's equivalence to the inherently iterative nature of Prim's algorithm for constructing an MST. Our algorithm performs a similar iterative step to perform the agglomerative labeling, but we perform this step on the sorted edges of an MST, after all needed distances have been computed. As we describe below, we use a novel variant of Borůvka's algorithm to construct the MST in parallel. In \autoref{sec:description}, we note that our formulation incresases the overall complexity to $O(N^2 + Nk\log N)$ but enables a high amount of parallelism.

Several approaches exist to perform single-linkage hierarchical clustering in parallel architectures \cite{olson1995parallel}. Many are variants of Borůvka and often combine the affinities graph construction with the MST by computing distance metrics right in the solver \cite{arefin2012knnagglom,Raff_2017}. Other approaches build upon Kruskal's algorithm, either by exploiting parallelism within different steps, such as the sorting step, or by building many trees independently in parallel and merging them into a single hierarchy \cite{hendrix2013scalable,hendrix2012parallel}. The latter approach forms the basis for an end-to-end parallel algorithm for single-linkage hierarchical clustering, even up to the dendrogram construction, but at the cost of performing many redundant computations and duplicating memory. While there have been claims that only the pairwise distance computations can benefit from GPU-acceleration \cite{chang2009hierarchical}, we demonstrate in \autoref{sec:benchmarks} that the linkage and cluster extraction steps also find performance gains.

The FAISS library is well known for containing state-of-the-art implementations of both exact and approximate nearest neighbors search on the GPU \cite{faiss}, though their exact implementation of nearest neighbors, known as brute-force, computes and stores an intermediate buffer of pairwise distances, performing a k-selection on each buffer. As we highlight in \autoref{sec:description}, we improve upon this design by fusing the computation of the distances with the k-selection when $k \leq 64$, allowing us to lower the memory footprint while also reducing the number of reads and writes to global memory.

Fast $k$-NN computations are critical for many types of algorithms in a class we refer to as 'neighborhood methods' which include information retrieval, clustering, dimensionality reduction, and classification/regression. Centroid-based clustering algorithms such as $k$-means and $k$-medioids rely on a fast computation of $1$-nearest neighbors, or closest centroid, to each training data point. Manifold learning algorithms such as TSNE \cite{tsne_cuda} and UMAP \cite{umap_gpu} rely on a special class of worse-case $k$-NN computations known as all-neighbors problems \cite{all_neighbors_knn,sankaranarayanan2007fast} to construct a graph of affinities. Similar to single-linkage clustering, the HDBSCAN algorithm, which can be formulated as a special case of single-linkage clustering, these problems specifically require a fast $k$-NN as all $n^2$ point pairs (or $\frac{n}{2}$ in metric spaces \cite{shalom2009hierarchical}) might need to be computed in the worst case for exact results. 

A natural optimization for $k$-NN-based approaches is to reduce computations by shrinking dimensionality \cite{pan2011fast}, partitioning \cite{shalom2011efficient,dash2007ppop,cayton2012accelerating} and/or quantizing \cite{jegou2010product} the space. Such methods can offer further speedups, sometimes at the expense of correctness. cuSLINK demonstrates state-of-the-art performance, but the primitives-based approach also leaves room for more optimized $k$-NN algorithms to be used.

Implementation of the Borůvka algorithm on the GPU has been considered before by several authors \cite{DaSilvaSousa2015,Vineet2009,Arefin2012}. Early attempts demonstrated speedups in comparison to parallel CPU solutions but focused on optimizing specific sparsity patterns \cite{DaSilvaSousa2015} or data structures \cite{Harish2009}. Those results worked for the initial goal, but tended to break down as the scope changed \cite{Arefin2012}. The breadth of MST applications kept growing meanwhile \cite{McInnes2017}, and translated to many distribution patterns and use cases. Designing a fast parallel solution resilient to multiple graph types and properties remains a primary concern.

Recent shifts in human communication, such as mobile phones and the internet of things (IoT), coupled with technical progress, has triggered an extensive growth of data volume. Scalability and speed have become a major concern for MST solvers. Previous solutions tended to primarily focus on performance, often at the cost of generous memory allocations. One of the strengths of the GPU architecture is the memory bandwidth, but the memory size is limited. We designed our MST algorithm to be memory efficient by avoiding explicit graph coarsening in order to scale further than previous implementations: past a billion edges on a single GPU.

A common problem in previous variants is cycle formation. Multiple edges of equal minimum weight between two components lead to multiple equivalent solutions. This results in ties when considering parallel execution, and can be a source of non-deterministic outputs. Weight alterations have been successfully applied to algebraic multi-grid aggregation in the past as a way to extract parallelism and break ties between strongly coupled nodes \cite{Naumov2015}. Unfortunately, random weight alterations cannot be applied directly without risking a change in the global relative ordering of the weights and thus the MST solution. In this paper, we propose an alteration that guarantees that all weights are different while preserving their relative order.

Our implementation adopts the \textit{scipy.hierarchy.linkage} format, which is also used by Scikit-learn, making it directly available on a trained \textit{AgglomerativeClustering} estimator. In addition to enabling SLINK, our MST implementation is also capable of constructing a maximum spanning tree, which enables our hierarchical clustering algorithm to also compute a complete-linkage clustering.

\section{cuSLINK}
\label{sec:description}

Our strategy to implement a faster SLINK is outlined in Algorithm \autoref{alg:compute_slhc}, where each line denotes a sub-step with references to the section where each step is detailed further. Current SLINK methods do significant unnecessary work by computing the entire pairwise adjacency matrix at once, but maximize compute --- or perform smaller work chunks but lose compute efficiency. Our strategy will instead sub-divide the work into chunks that perform some redundant work, but allow sufficient compute efficiency to result in a significant net speedup.

\begin{algorithm}[!h]
  \caption{cuSLINK Algorithm}
  \label{alg:compute_slhc}
  \begin{algorithmic}[1]
    \State \textbf{Input:} X, n\_clusters
    \State \textbf{Output:} 1-d array of cluster labels
    \State {knn\_graph = compute\_connectivities(X)} \Comment{\autoref{sec:knn_graph_construction}, Alg. \autoref{alg:fused_knn}}
    \State {mst\_edges, colors = mst(knn\_graph)} \Comment{\autoref{sec:spanning_trees}, Alg. \autoref{alg:pmst}}
    \State {mst\_edges = connect\_graph(knn\_graph, mst\_edges, colors)} \Comment{\autoref{sec:knn_graph_construction}, Alg. \autoref{alg:connect_graph}}
    \State {dendrogram = relabel(mst\_edges)} \Comment{\autoref{sec:relabel_into_dendrogram}}
    \State {return extract\_clusters(dendrogram, n\_clusters)} \Comment{\autoref{sec:cut_the_dendrogram}, Alg. \autoref{alg:extract_flattened_clusters}}
\end{algorithmic}
\end{algorithm}

We start by converting tabular inputs from $N$ points in $d$ dimensions into a sparse $k$-NN graph of affinities, which is then used as input to compute the MST. As mentioned, the original SLINK algorithm computes the MST sequentially by computing and maintaining the $1$-NN of the data points (and clusters) while constructing each of the O($\log N$) levels of the resulting dendrogram. Our cuSLINK breaks the algorithm into four steps: 1) construction of affinities graph, 2) construction of MST, 3) construction of the dendrogram, and 4) extraction of flat cluster assignments from the dendrogram. This separation uses Amdahl's Law to allow parallelism of the most performance-critical pieces and isolates the sequential bits to only the third step.

In order for an MST to converge, the edges that it has available as input need to form a connected graph and it is not guaranteed that the closest $k$ neighbors to each data point will form a connected graph. In the case where the MST construction doesn't converge, resulting in a minimium spanning forest (MSF, i.e., multiple MSTs that are not connected), we compute an additional $1$-NN across the resulting super-vertexes in the MSF and perform another iteration of the MST with the new edges to connect the super vertices. It's possible there could be multiple iterations of computing $1$-NNs until the MSF converges to an MST, the number of iterations bounded from above by O($\log N$) in the worst case when $k=1$. The algorithm for connecting anotherwise disconnected $k$-NN graph is outlined in Algorithm \autoref{alg:connect_graph}

\begin{algorithm}
  \caption{connect\_graph()}
  \label{alg:connect_graph}
  \begin{algorithmic}[1]
    \State \textbf{Input:} knn\_graph, mst\_edges, colors
    \State \textbf{Output:} additional mst\_edges to connect knn\_graph
    \While{{n\_unique(colors) $\neq$ 1}}
     \State {mst\_edges = mst\_edges $\cup$  cross\_color\_1nn(knn\_graph, colors)} \Comment{Alg. \autoref{alg:fused_1nn}}
     
     \State {mst\_edges, colors = mst(mst\_edges)} \Comment{Alg. \autoref{alg:pmst}}
    \EndWhile    
    \State {return mst\_edges}
\end{algorithmic}
\end{algorithm}

The $O(E\log V)$ complexity of computing the MST itself becomes $O(Nk\log N)$ in our formulation, since $k$-NN bounds the number of edges, $E$, to $N \times k$ and $V=N$. Our implementation uses brute-force $k$-NN, so we still perform $N^2$ distance computations, resulting in $O(N^2 + Nk\log N)$ overall commputational complexity. While this complexity is slightly worse than the SLINK algorithm, it does so because it allows more parallel computation to occur with less overhead, resulting in a lower total runtime.

As we describe in Section \autoref{sec:cut_the_dendrogram}, flattened cluster assignments are obtained for each point $n \in N$ by cutting the dendrogram at the level which yields the desired number of clusters ands assigning a unique label to each resulting connected component. While not currently supported by cuSLINK, complete-linkage clustering can be obtained by computing the maximum spanning tree and sorting the resulting edges in descending order. With our building blocks, it's also possible to implement a variant which accepts a distance threshold for which to cut the dendrogram, however we leave this to future work.  

Unlike the original SLINK formulation, cuSLINK separates the construction of the dendrogram from the $k$-NN graph and MST steps in order to maximize parallelism of the former steps since the order of dendrogram construction is important, making it inherently sequential. Algorithm \ref{alg:compute_slhc} shows the cuSLINK steps and Algorithm \ref{alg:connect_graph} shows the steps to finding additional edges for connecting the super-vertices.

\subsection{Nearest Neighbors}
\label{Hierarchical Clustering}

\subsubsection{$k$-NN Graph Construction}
\label{sec:knn_graph_construction}

The connectivities graph is constructed in parallel using a novel GPU-accelerated brute-force $k$-NN primitive that, as we will outline below, is able to fuse the k-selection steps with computation of the distances in order to lower the required memory footprint and remove the need for additional steps downstream.

A common approach to computing a brute-force $k$-NN is to first break up the computation of the full $m \times n$ pairwise distance matrix into smaller tiles, each requiring an intermediate buffer of device memory. A k-selection is performed as a follow-on computation over each tile of pairwise distances to reduce the columns down to the $k$ smallest (or largest) distances and output them to another memory buffer. If tiling over both dimensions of the pairwise distance matrix, additional k-selection steps might be needed. These intermediate tiles and k-selection steps require additional memory accesses that can be avoided when $k$ is small enough such that the closest neighbors can be reduced into shared memory from each warp and selected within each thread-block. We use this novel insight to develop a fast $k$-NN strategy outlined in Alg. \autoref{alg:fused_knn}.

\begin{algorithm}[!h]
  \caption{Fused $k$-NN}
  \label{alg:fused_knn}
  \begin{algorithmic}[1]
    \State {gridStrideY = curBlockY $\times$ batchM}    \Comment{Determine output tile coordinates}
    \State {\_\_shared\_\_ kvp\_t topKs[batchM , 64]} \Comment{Allocate shared mem top-k store}
    
    \For{{gridStrideY < m}; \textbf{step} {batchM * nBlocksY};}
     \State {gridStrideX = curBlockX $\times$  
 batchN}
     \State{init\_smem(topKs)} \Comment{Initialize shared mem}
         \For{{gridStrideX < n}; \textbf{step} {batchN $\times$ nBlocksX};}
            \State{prolog(gridStrideX, gridStrideY)} \Comment{Zero init shared mem accumulator}
            \State{gemm\_nt\_loop()} \Comment{Compute dot product along k}
            \State{epilog\_op\_topk(gridStrideX, gridStrideY)} \Comment{Norm and top-k, Alg. \autoref{alg:epilog_op_top_k}}
         \EndFor
     \State{row\_epilog\_op\_topk(gridStrideY)} \Comment{Store topk in global mem, Alg. \autoref{alg:row_epilog_op_top_k}} 
    \EndFor   
\end{algorithmic}
\end{algorithm}

For the case where $k\leq64$, special optimizations can be done to the k-selection in the GPU implementation of the FAISS' \cite{faiss} k-selection primitive. Our improved version is detailed in Alg. \autoref{alg:row_epilog_op_top_k} in \autoref{sec:additional_algos}. This step imposes a lot of pressure on the registers as each warp maintains its own thread-level queue of new values to be merged along with a warp-wide queue of the fully sorted top-k values and a register to store the current largest value in the warp-wide queue for early filtering of candidates. When the thread-level queues reach their max for new potential top-k values, a sorting and merging step is initiated synchronously in the warp by performing a bitonic sort of the thread-level values with the warp-level queue, reducing the warp-level queue back down to size $k$. Our warp-level k-selection routine follows the same general design used in FAISS, but reduces the number of warp-selection computations, which require expensive synchronization between threads within each warp, using the stream compaction technique \cite{bakunas2017efficient} within each block.

At the block level, each warp performs its warp-level k-selection and stores the resulting $k$ selected distances and indices in shared memory. Next, in a grid-stride, each thread discards all distances and indices which are less than the previous warp-level k-selection, performing the stream compaction to write the new set of filtered indices and distances to shared memory. A warp-level \textit{ballot\_sync()} is performed over the filtered counts, $2 \times 256$ per warp and $2 \times 8$ per thread, resorting to a scan only if the number of filtered counts is $>0$. For each set of $2 \times 256$ distances computed by a warp, the warp-select calls \textit{any\_sync()} a total of 128 times per grid-stride by whole thread block. With stream compaction, \textit{ballot\_sync()} needs to be called only 16 times in the best case.

\subsubsection{$k$-NN Graph Connection}
\label{sec:knn_graph_connection}

Unfortunately, a $k$-NN alone with only a naive choice of $k$ doesn't often scale to larger datasets in practice as the size of $k$ needed for the MST to converge to a single supervertex can grow significantly large, being bounded above by $k=N$ in the worst-case. Since our MST implementation is robust to this type of input, a minimum spanning forest (MSF) will be returned when this occurs and additional edges are added to the MSF by performing a $1$-NN query that connects points only across different supervertices, or components, together before re-computing the MST.

\begin{algorithm}
  \caption{Fused $1$-NN}
  \label{alg:fused_1nn}
  \begin{algorithmic}[1]
    \State {gridStrideY = curBlockY $\times$ batchM}    \Comment{Determine output tile coordinates}
    \State {min\_kvp = (MAX\_FLOAT, MAX\_INT)} \Comment{Init min with max values}
    
    \For{{gridStrideY < m}; \textbf{step} {batchM * nBlocksY};}
     \State {gridStrideX = curBlockX $\times$  
 batchN}
         \For{{gridStrideX < n}; \textbf{step} {batchN $\times$ nBlocksX};}
            \State{prolog(gridStrideX, gridStrideY)} \Comment{Zero init shared mem accumulator}
            \State{gemm\_nt\_loop()} \Comment{Compute dot product along k}
            \State{epilog(gridStrideX, gridStrideY)} \Comment{L2 norm addition}
         \EndFor
     \State{row\_epilog\_op(gridStrideY)} \Comment{Global mem min reduce across all tiles for row} 
    \EndFor   
\end{algorithmic}
\end{algorithm}

Similar to the fused $k$-NN primitive outlined above, Algorithm \autoref{alg:fused_1nn} shows the general steps for our novel $1$-NN primitive for Euclidean-based distances, which also fuses together the computation of the minimum neighbor with the distance computations. Since $k=1$, we can avoid the sorting and merging of the warp-selection altogether by computing and storing a single $min$ as the distances are computed. This effectively enables the use of registers alone within each thread of the GPU for fast storage, comparison, and computation of the closest neighbor.

\subsection{Spanning Trees}
\label{sec:spanning_trees}

Since the graph is assumed to be undirected ($w_{i,j} = w_{j,i}$), the resulting adjacency matrix is symmetric. Our algorithm is resilient to negative weights and can solve the maximum spanning tree problem natively. The maximum spanning tree has weights greater than or equal to the weight of every other spanning tree, and is found by forming the additive inverse $G'=(V,-E)$ and solving the MST problem on $G'$.

Algorithm \ref{alg:pmst} provides an overview of our parallel MST implementation which relies on the property that the minimum incident edge, or closest neighbor to each vertex, has to be in the MST (e.g., same as original SLINK). The first step is to identify these edges. Then, edge contraction of the minimum incident edges is applied recursively until a steady state is reached. Instead of explicitly contracting edges through expensive graph coarsening, the MST components are represented using labels (i.e. colors). The problem is reduced to finding the minimum incident edge across color boundaries \cite{Vineet2009}.

Our MST solution is artificially made unique by applying a weight alteration \autoref{weight_alteration}. MST and MSF are almost identical problems with the difference that MSF refers to the case where $G$ has multiple strongly connected components. The latter is particularly relevant for $k$-NN applications which may consist of multiple connected components described in \autoref{Hierarchical Clustering}. The historical Borůvka MST formula iterates until only one color, or super-vertex, remains. Instead, to find an MSF we detect a steady state and exit if no minimum incident edge to another color has been found.

\begin{algorithm}
  \caption{MST Algorithm}
  \begin{algorithmic}[1]
    \State \textbf{Input:} G
    \State \textbf{Output:} mst\_edges, colors
    \State {G\_altered = weight\_alteration(G) \Comment{\autoref{weight_alteration}}} 
    \While{{!exit}}
     \State {min\_edges = min\_edge\_per\_vertex(G\_altered, colors) 
     \State {new\_mst\_edges = min\_edge\_per\_supervertex(min\_edges)}
     \Comment{\autoref{min_edge}}}
     \State{exit = len(new\_mst\_edges)}
     \If{{exit}} \Comment{Return MSF by exiting}
        \State{break}
     \EndIf
     \State{done = false} \Comment{Continue iterating over label\_propagation when necessary}
	 \State {new\_colors = label\_propagation(new\_mst\_edges, done)} \Comment{\autoref{label_prop}}
    \EndWhile
    \State{return new\_mst\_edges, new\_colors}
  \end{algorithmic}
  \label{alg:pmst}
\end{algorithm}

We leverage the Compressed Sparse Row (CSR) format as input to our MST primitive because it enables an efficient memory access pattern.

\subsubsection{Weight Alteration} \label{weight_alteration}
Cycle detection and removal was identified as one of the main challenges in previously published GPU solutions \cite{Vineet2009}. The problem can be reduced to selecting an incident edge to every vertex in parallel without creating cycles. Multiple edges of equal minimum weight between two components result in multiple equivalent solutions for the MST problem. In parallel, each component could select a different edge which would result in adding a cycle between them and breaking the tree structure. To address this issue, we propose a simple solution to prevent cycle formation by generating an alteration on the edge weights that guarantees that all weights are different while preserving the relative order of all weights. For any graph that has only a distinct set of edges, this produces a deterministic result. As a result, all vertices can consistently select edges in an embarrassingly parallel fashion. Notice that this has the side effect of altering explicit zeroes, which means our solution cannot support graphs that have zeroed edge weights. The technique for altering the weights is done in 3 steps:

\begin{itemize}
\item Identify $\theta>0$, where $\theta$ is the minimum edge weight difference between any two pairs of edges in the graph
\item For each edge in the upper triangular side of the matrix, add noise to each edge weight by generating a random number $\epsilon \in [0, \theta)$.
\item Replicate the upper triangular part into the lower triangular part so that $w_{ij} = w_{ji}$\\
\end{itemize}

\subsubsection{Minimum incident edges} \label{min_edge}

A common solution, as noted with \cite{Harish2009} and \cite{DaSilvaSousa2015}, is to assign one thread per vertex to scan all edges. However, this comes with the risk that the kernel becomes bound to the slowest performing thread. With the help of \autoref{weight_alteration}, we overcome the issue of tie-breaking when trying to find the minimum outgoing edge per supervertex as each edge has a unique weight. Thus, overcoming the cycle-detection problem, we divide the task of finding the minimum outgoing edge in two sub-tasks:
\begin{itemize}
    \item \emph{Minimum Edge Per Vertex}: Using the CSR format, we assign 32 threads (one warp) to scan the edges of each vertex. These threads, using shared memory reduction, find an outgoing edge to a destination vertex that is part of a different supervertex than the source and also atomically record a minimum edge weight for that source supervertex
    \item \emph{Minimum Edge Per Supervertex} : Continuing from the previous task where we found the minimum outgoing edge for each vertex, we still need to reconcile the minimum source, destination pair for each supervertex. Whichever vertex found an edge that was the minimum for its supervertex adds it to the final solution
\end{itemize}

Note that self edges are automatically invalid because they point to the source's color.

\subsubsection{Label Propagation} \label{label_prop}
Label propagation in \cite{Vineet2009} is performed by ensuring all vertices explicitly try to converge to supervertex 0, while \cite{DaSilvaSousa2015} uses graph contraction to form a new graph of supervertices in the current MST iteration. Instead, our label propagation improves upon the speed of the former and memory requirements of that latter by indirectly keeping track of supervertices using the \textit{color} and \textit{supervertex} arrays. By working only on newly added MST edges in a given iteration, we ensure that fewer iterations of label propagation are needed compared to preceding MST iterations as each iteration adds fewer edges to the solution. This was instrumental in removing the bottleneck that comes with wide graphs (such as road networks). Initially, each \textit{$v_i$} holds the color \textit{i}.
\linebreak
Alg. \ref{alg:mincol} shows  how the minimum colors between source and destination vertices is resolved, for each newly added edge to the MST.
\begin{algorithm}
  \caption{min\_pair\_colors($V$)}
  \label{alg:mincol}
  \begin{algorithmic}[1]
    \For{$vertex \in V$} \Comment{The incident vertex}
        \State{edge = filtered\_min\_edges[vertex]}
        \If{found\_by\_vertices(edge)}
            \State{neighbor\_vertex = get\_neighbor\_vertex[edge]}
            \Comment{The neighbor vertex this edge connects}
            \State{supervertex = supervertices[vertex]}
            \State{color = colors[supervertex]}
            \State{neighbor\_supervertex = supervertices[neighbor\_vertex]} 
            \State{neighbor\_color = colors[neighbor\_supervertex]}
            \State{atomicMin(next\_color[supervertex], neighbor\_color)}
            \State{atomicMin(next\_color[neighbor\_supervertex], color)}
        \EndIf
    \EndFor
  \end{algorithmic}
\end{algorithm}
\linebreak

The color for every $v \in V$ whose supervertex changed colors in Alg. \ref{alg:mincol} gets updated in Alg. \ref{alg:mincol2}. We determine whether Alg. \ref{alg:mincol} and Alg. \ref{alg:mincol2} need to be iterated on again, in case a supervertex has not reached its final color as defined by newly added MST edges.

\begin{algorithm}
  \caption{update\_colors($V$)}
  \label{alg:mincol2}
  \begin{algorithmic}[1]
    \For{$vertex \in V$}
        \State{color = colors[vertex]} 
        \State{supervertex = supervertices[vertex]}
        \State{new\_color = next\_color[supervertex]} 
          \If{color > new\_color}        
            \State {colors[vertex] = new\_color}
            \State {done = false}
          \EndIf
    \EndFor
  \end{algorithmic}
\end{algorithm}

Finally, we propagate and resolve colors for the entire topology by updating vertices whose supervertices changed colors.

\subsection{Dendrogram}
\subsubsection{Relabel into Dendrogram}
\label{sec:relabel_into_dendrogram}

After computing the minimum spanning tree on the connectivities graph, the resulting $N-1$ edges are sorted in parallel by weight. The dendrogram is constructed on the CPU by expanding the total set of vertices from $N$ to $(N-1) * 2$ and renumbering the original vertices as they are merged together into the hierarchy. A parent vertex for any level $i$ in the hierarchy, where $0 \le i < \frac{N-1}{2}$, will always be $\ge N$ and can be computed with the simple formula $i + N$. A union-find structure with union-by-rank and path compression is used to achieve runtime of $O(N\alpha(N))$ \cite{McInnes2017,tarjan1975efficiency}.

The strict ordering of the dendrogram construction step makes it inherently sequential \cite{chang2009hierarchical,olson1995parallel}. More recently, the optimization outlined by \cite{McInnes2017} produces acceptable performance on the CPU, so we exploit as much parallelism in the remaining steps as possible.

\subsubsection{Cut the Dendrogram}
\label{sec:cut_the_dendrogram}

\begin{algorithm}
  \caption{extract\_flattened\_clusters(dendrogram)}
  \label{alg:extract_flattened_clusters}
  \begin{algorithmic}[1]
  \State{label\_roots = find\_label\_roots()} \Comment{Find the root nodes for each label}
    \State {sort(label\_roots)}
    \State{cut\_level = (n\_points -1) - (n\_clusters - 1)}
    \State {labels = inherit\_labels(cut\_level, dendrogram)} \Comment{Leaves inherit from label roots}

\end{algorithmic}
\end{algorithm}

\begin{algorithm}
  \caption{inherit\_labels(cut\_level, dendrogram)}
  \label{alg:inherit_labels}
  \begin{algorithmic}[1]
  \For{$vertex \in dendrogram$} \Comment{Loop through all nodes in dendrogram}
    \State {cur\_level = get\_tree\_level(vertex)}
    \If{cur\_level $\le$ cut\_level}
    `\State{cur\_label = get\_parent\_label()}
    \While{!is\_labeled(cur\_label)} \Comment{Iterate parents until label is found}
        \State{cur\_parent = get\_parent()}
        \State{cur\_level = get\_parent\_tree\_level()}
        \State{cur\_label = get\_parent\_label()}
    \EndWhile
    
    \State {label[vertex] = cur\_label} \Comment{Assign label of labeled parent}
    \EndIf
  \EndFor

\end{algorithmic}
\end{algorithm}

Cluster assignments (Alg. \autoref{alg:extract_flattened_clusters}) are extracted from the dendrogram by first cutting it at a particular level, yielding a desired $n\_clusters$ number of cluster tree roots. These tree roots are computed by sorting the last $n\_clusters * 2$ elements of the dendrogram and extracting the smallest $n\_clusters$ elements from the sorted array. Unique labels are given to each of these tree roots and all nodes in levels of the tree lower than the dendrogram label root nodes inherit the labels from their closest labeled ancestors in parallel (Alg. \autoref{alg:inherit_labels}).

\section{Experiments}
\label{sec:benchmarks}

In this section, we benchmark both cuSLINK end-to-end along with our $k$-NN, $1$-NN and MST primitives. We show that all components of cuSLINK outperform the available state-of-the-art solutions. All benchmarks are performed w/ CUDA 11.8 on Nvidia A100 GPUs. We selected our comparison implementations based on availability of packages and/or source code which was able to built and run by our best effort. We also made every effort to update existing state-of-the-art solutions to run CUDA 11.8 when needed. Note that we compare against implementations on the CPU only when corresponding GPU source code was not available.

\subsection{Nearest Neighbors}

We measured the performance of our fused $1$-NN and $k$-NN implementations against FAISS on the GPU, which is the current known state-of-the-art for k-selection and brute-force nearest neighbors on the GPU.

\begin{table}[!h]

\caption{Performance comparison between FAISS and cuSLINK's $k$-NN on randomly generated data for $k=32$. Our Fused $k$-NN enables consistent speedups at all sizes, particularly smaller sizes where memory transfers dominate all time spent. By fusing the operations we do not require additional allocation.}
\centering
\label{tab:fused_knn}
\begin{tabular}{ l  l  c  r  r }
\toprule
Index Rows & Query Rows &  GPU-FAISS & cuSLINK\\ \midrule
100K & 100K    & 261ms     & \textbf{143ms}\\ 
200K & 200K   & 783ms      & \textbf{537ms} \\
400K & 400K   & 2706ms & \textbf{2017ms}\\
1M & 1M   & 1.607s        & \textbf{1.218s} \\
\bottomrule
\end{tabular}
\end{table}

\begin{table}[!h]
\caption{Performance comparison between FAISS and cuSLINK's $1$-NN on randomly generated data. Similar to \autoref{tab:fused_knn} our performance dominates, even more dramatically for lower sizes --- up to 178$\times$ faster. Due to the iterative nature of \autoref{alg:compute_slhc} and the merging in HAC, problems of all sizes will occur during a larger clustering, so all performance levels are relevant to final speedup.}
\centering
\label{tab:fused_1nn}
\begin{tabular}{ l  l  l  l  l }
\toprule
Index Rows & Query Rows & Cols & GPU-FAISS & cuSLINK\\ \midrule
100K & 100 & 128   & 98.4ms     & \textbf{0.55ms}\\ 
100K & 100 & 256  & 95.6ms      & \textbf{0.967ms} \\
100K & 1k & 64  & 96.6ms & \textbf{1.85ms}\\
100K & 1K & 128   & 98.9ms     & \textbf{3.39ms}\\ 
100K & 1K & 256  & 104ms      & \textbf{6.46ms} \\
100K & 10K & 64  & 126ms & \textbf{17ms}\\
100K & 10K & 128   & 146ms     & \textbf{32ms}\\ 
100K & 10K & 256  & 156ms      & \textbf{62.2ms} \\\bottomrule
\end{tabular}

\end{table}

\subsection{Spanning Tree}
Previous work already showed that parallel MST solver on GPU outperformed CPU versions \cite{Vineet2009}. Hence we compare against previous GPU implementations and consider CPU comparisons to be out of the scope of this paper. While performance of \cite{Harish2009} is better than \cite{Vineet2009}, we did not compare against it because the bitwise technique greatly limits the supported input size, as shown in \cite{Arefin2012}.

In Table \ref{tab:graphs} we selected road networks from the 9th DIMACS challenge to compare against the experiments performed by \cite{Vineet2009}.

\begin{table}[!h]
\caption{We selected road networks from the 9th DIMACS challenge to compare against the experiments performed by \cite{DaSilvaSousa2015}. Compared to the prior state-of-the-art algorithm, our method is always faster and up to 3.5$\times$ faster as the problem size increases.}
\centering
\label{tab:graphs}
 \begin{tabular}{ l  r  r  r  r }
\toprule
Description & no. nodes & no. edges & Sousa2015 & cuSLINK\\ \midrule
New York City & 263,346 & 733,846 & 29.265ms & \textbf{20.217ms}\\ 
SF Bay Area & 321,270 & 800,172 & 32.689ms & \textbf{22.606ms}\\
CO & 435,666 &  1,057,066 & 38.819ms & \textbf{23.680ms}\\
FL & 1,070,376 & 2,712,7986 & 82.822ms & \textbf{35.552ms}\\
Northwest USA &  1,207,945 &  2,840,208 & 84.203ms & \textbf{36.884ms}\\ 
Northeast USA & 1,524,453 & 3,897,636 & 112.173ms & \textbf{51.879ms}\\
CA \& NV &  1,890,815 & 4,657,742 & 132.726ms & \textbf{61.366ms}\\
Great Lakes & 2,758,119 &  6,885,658 & 191.827ms & \textbf{81.994ms}\\
Eastern USA & 3,598,623 &  8,778,114 & 265.426ms & \textbf{96.100ms}\\
Western USA & 6,262,104 & 15,248,146 & 450.833ms & \textbf{127.545ms}\\
Central USA & 14,081,816 & 34,292,496 & 1004.624ms & \textbf{278.841ms}\\
Full USA & 23,947,347 & 58,333,344 & 1685.172ms & \textbf{478.898ms}\\
\bottomrule
\end{tabular}

\end{table}

On larger road networks, our implementation scales better than \cite{DaSilvaSousa2015}. Recall that the latter is explicitly forming super-vertices which becomes increasingly expensive as the size of the problem increases.

\subsection{Single-linkage Hierarchical Agglomerative Clustering}
Since dendrogram construction is not often exposed as an independent step, we evaluate our end-to-end single-linkage hierarchical clustering implementation on the GPU against Scikit-learn's \textit{AgglomerativeClustering} implementation on the CPU in Table  \ref{tab:slhcsklearn}. Each experiment was performed on a NVIDIA DGX1 using several real-world datasets, often encountered in clustering and nearest neighbors research.
These experiments demonstrate that the performance of our SLHC implementation has the potential to lower the time spent in compute during the data analysis process, enabling near real-time speeds for data sets that take nearly 40 minutes to process on the CPU.

\begin{table}[!h]
\caption{End-to-end execution times comparing our GPU-accelerated SLHC implementation against Scikit-learn on the CPU for real-world datasets. In most cases Scikit-learn times out (after 24 hours), and so no result is available.}
\centering
\label{tab:slhcsklearn}
\begin{tabular}{@{}lcccc@{}}
\toprule
Dataset & Shape & Clusters & Scikit-learn & cuSLINK\\ \midrule
Deep-1B  & 8M$\times$96   & 100   & ---     & \textbf{1806s}\\ 
SIFT-128 & 1M$\times$128  & 100     & ---      & \textbf{37.23s} \\
NYTimes & 290k$\times$256 & 100 & --- & \textbf{9.227s}\\
MNIST &  60K$\times$784  & 10      & 2171s        & \textbf{0.926s} \\
Fashion MNIST & 60K$\times$784 & 25        & 2169s       & \textbf{0.947s}\\ 
\bottomrule
\end{tabular}
\end{table}

\section{Conclusion}

In addition to a novel, modular, and state-of-the-art implementation of single-linkage hierarchical clustering, we've outlined and contributed multiple reusable, novel and state-of-the-art primitives in this paper.
These primitives include k-nearest graph construction, graph-based minimum spanning tree solver, and a novel parallel dendrogram cluster extraction method.

We demonstrated that our primitives are both flexible and fast, and have a potential to impact several different industries as existing state-of-the-art methods for end-to-end single-linkage clustering are intractable on large datasets. These primitives are all fully open source and available as part of the RAFT library (https://github.com/rapidsai/raft). 
\bibliographystyle{splncs04}
\bibliography{mst_references}

\appendix

\renewcommand{\thesection}{Appendix \Alph{section}}  \renewcommand{\thesubsection}{\Alph{section}.\arabic{subsection}}

\section{TopK Epilog Operations}
\label{sec:additional_algos}
\begin{algorithm}[H]
  \caption{row\_epilog\_op\_topk(gridStrideY)}
  \label{alg:row_epilog_op_top_k}
  \begin{algorithmic}[1]
    \If { nBlocksX == 1}    
    \State {loadAllWarpQShmem()} \Comment{Load final warpQs of all the warps from shared mem}
     \State {storeWarpQGmem()} \Comment{Store it in global memory}
 \Else
     \State{acquire\_gmem\_mutex()} \Comment{ Acquire mutex for this block of rows}
     \State{merge\_topks\_into\_gmem()} \Comment{Merge its topKs with global mem}
    \State{release\_gmem\_mutex()} \Comment{Release mutex for this block of rows}
\EndIf
\end{algorithmic}
\end{algorithm}

\begin{algorithm}[H]
  \caption{epilog\_op\_topk(gridStrideX, gridStrideY)}
  \label{alg:epilog_op_top_k}
  \begin{algorithmic}[1]
    \If {first\_tile\_of\_row}    
    \State {perform\_warp\_select()} \Comment{Filter potential top-k and sort in chunks of 64}
    
    \Else
    
     \State {load\_prev\_topks()} \Comment{Load previous topks from shared mem to registers}
     \State{anyWarpTopKs = check\_any\_potential\_topks()} \Comment{Check chunk for topKs}
     \State{anyWarpTopKs = \_\_syncthreads\_or(anyWarpTopKs > 0)} \Comment{Check warps}
    \If{anyWarpTopKs}
    \State{stream\_compaction\_to\_smem()} \Comment{Store potential topks in shared mem}
    \State{\_\_syncwarp();} \Comment{Norm and top-k}
 \State{perform\_warp\_select()} \Comment{Warp select or insert into warpQ} 
     \State{storeWarpQShmem()} \Comment{Final topks from warpQ registers to shared mem }
     \EndIf
\EndIf
\end{algorithmic}
\end{algorithm}

\section{GPU Architecture}
\label{sec:gpu_architecture}

The largest GPUs today contain thousands of hardware processing cores that are divided into groups of 32 or 64, which can execute instructions concurrently. The instructions for each group of cores are orchestrated by a streaming multiprocessor (SM). From a programming perspective, these groups are called warps and each physical core executing an instruction is known as a thread. Each warp can process a single instruction at a time across its threads in parallel using a paradigm called single-instruction multiple data (SIMD). 

It's important that threads within a warp minimize conditional branching that will cause the threads to wait for each branch to complete before proceeding. This is called thread divergence, and can severely limit effective parallel execution. On the Volta and Ampere architectures, each SM can track the progress of up to 64 warps concurrently \cite{tesla2018v100}, and rapidly switch between them to fully utilize the SM. Each SM has a set of registers available which allows warps to perform collective operations, such as reductions. Warps can be grouped into blocks and a small amount of memory can be shared across the threads and warps. 

Global, or device, memory can be accessed by all of the SMs in the GPU. Accesses to contiguous device memory locations within a warp can be coalesced into a single blocked transaction so long as the accesses are performed in the same operation. In SIMD architectures, uniform patterns can be critical to performance unless latencies from non-uniform processing, such as uncoalesced memory accesses, can be hidden with increased parallelism.

Registers provide the fastest storage, and it's generally preferable to perform reductions and arithmetic as intra-warp collective operations where possible. Intra-block shared memory is also generally preferred over global memory when a problem can be made small enough to benefit.

\section{Code Examples}

\begin{figure}[!h]
 \begin{minipage}{0.991\columnwidth}
\begin{minted}[breaklines]{c++}
#include <raft/core/device_mdarray.hpp>
#include <raft/core/device_resources.hpp>
#include <raft/distance/distance_types.hpp>
#include <raft/neighbors/brute_force.cuh>

uint32_t n_index_rows = 1000;
uint32_t n_query_rows = 100;
uint32_t n_features = 512;
uint32_t k = 5;

auto metric = raft::distance::DistanceType::L2Expanded;

raft::device_resources handle;
auto index = raft::make_device_matrix<float>(
    handle, n_index_rows, n_features);
    
auto query = raft::make_device_matrix<int64_t>(
    handle, n_query_rows, n_features);

auto out_indices  = raft::make_device_matrix<float>(
    handle, n_query_rows, k);
    
auto out_distsances = raft::make_device_matrix<int64_t>(
    handle, n_query_rows, k);

// .. Populate index and query with data ...

raft::neighbors::fused_l2_knn(
    handle, raft::make_const(index.view()),
    raft::make_const(query.view()),
    out_distances.view(), out_indices.view(),
    metric);

\end{minted}
 \end{minipage}
 \caption{C++ Code example of using the fused $k$-NN primitive API}
  \label{lst:fused_knn_example}
\end{figure}

\begin{figure}[!h]
 \begin{minipage}{0.991\columnwidth}
\begin{minted}[breaklines]{c++}
#include <raft/core/device_mdarray.hpp>
#include <raft/core/kvp.hpp>
#include <raft/core/device_resources.hpp>
#include <raft/distance/distance_types.hpp>
#include <raft/distance/fused_l2_nn.cuh>

uint32_t n_index_rows = 1000;
uint32_t n_query_rows = 100;
uint32_t n_features = 512;

using kvp_t = raft::KeyValuePair<uint32_t, float>;
auto metric = raft::distance::DistanceType::L2Expanded;

raft::device_resources handle;
auto index = raft::make_device_matrix<float>(
    handle, n_index_rows, n_features);
    
auto query = raft::make_device_matrix<uint32_t>(
    handle, n_query_rows, n_features);

auto out_kvp  = raft::make_device_vector<kvp_t>(
    handle, n_query_rows);
// ... Populate index and query with data ...

raft::neighbors::fused_l2_nn_min_reduce(
    handle, raft::make_const(index.view()),
    raft::make_const(query.view()), out_kvp.view(),
    metric);

\end{minted}
 \end{minipage}
 \caption{C++ Code example of using the fused $1$-NN primitive API}
  \label{lst:fused_knn_example}
\end{figure}

\begin{figure}[!h]
 \begin{minipage}{0.991\columnwidth}
\begin{minted}[breaklines]{c++}
#include <raft/core/device_mdarray.hpp>
#include <raft/core/device_resources.hpp>
#include <raft/cluster/single_linkage.cuh>

uint32_t n_rows = 1000;
uint32_t n_features = 512;
uint32_t n_clusters = 5;

auto metric = raft::distance::DistanceType::L2Expanded;

raft::device_resources handle;
auto X = raft::make_device_matrix<float>(
    handle, n_rows, n_features);
    
auto dendrogram = raft::make_device_matrix<float>(
    handle, n_rows-1, 2);
                                         
auto labels = raft::make_device_vector<uint32_t>(
    handle, n_rows);

// ... Populate X with data ...

raft::cluster::single_linkage(
    handle, raft::make_const(X.view()),
    dendrogram.view(), labels.view(),
    metric, n_clusters);

\end{minted}
 \end{minipage}
 \caption{C++ code example of using the single-linkage hierarchical clustering API}
  \label{lst:fused_knn_example}
\end{figure}

\begin{figure}[!h]
 \begin{minipage}{0.991\columnwidth}
\begin{minted}[breaklines]{c++}
#include <raft/core/device_csr_matrix.hpp>
#include <raft/core/device_coo_matrix.hpp>
#include <raft/core/device_resources.hpp>
#include <raft/sparse/solver/mst.cuh>

uint32_t n_rows = 1000;
uint32_t nnz = 20000;

raft::device_resources handle;
auto G = raft::make_device_csr_matrix<float>(
    handle, n_rows, n_rows, nnz);
    
auto mst_edges = raft::make_device_coo_matrix<float>(
    handle, n_rows, n_rows);

auto colors = raft::make_device_vector<int>(
    handle, n_rows);
// ... Populate G with data ...

raft::sparse::solver::mst(
    handle, raft::make_const(G.view()),
    mst_edges.view(), colors.view());

\end{minted}
 \end{minipage}
 \caption{C++ code example of using the minimum spanning tree API}
  \label{lst:mst_example}
\end{figure}

\end{document}